\documentclass[runningheads]{llncs}

\usepackage{eccv}

\usepackage{eccvabbrv}
\usepackage{wrapfig}
\usepackage{graphicx}
\usepackage{booktabs}

\usepackage[accsupp]{axessibility}  
\usepackage{multirow}
\usepackage{tablefootnote}

\usepackage{hyperref}

\usepackage{orcidlink}

\begin{document}

\title{MobileDiffusion: Instant Text-to-Image Generation on Mobile Devices} 

\titlerunning{MobileDiffusion}

\author{Yang Zhao \and
Yanwu Xu \and
Zhisheng Xiao \and Haolin Jia \and Tingbo Hou}

\authorrunning{Yang Zhao, Yanwu Xu, et al.}

\institute{Google \\
\email{yzhao63@buffalo.edu, yanwuxu@bu.edu\\
xiaozhisheng950@gmail.com, haolinmz@google.com, houtingbo@gmail.com}\\
}

\maketitle

\begin{figure}[!h]
    \centering
    \includegraphics[width=1.0\textwidth]{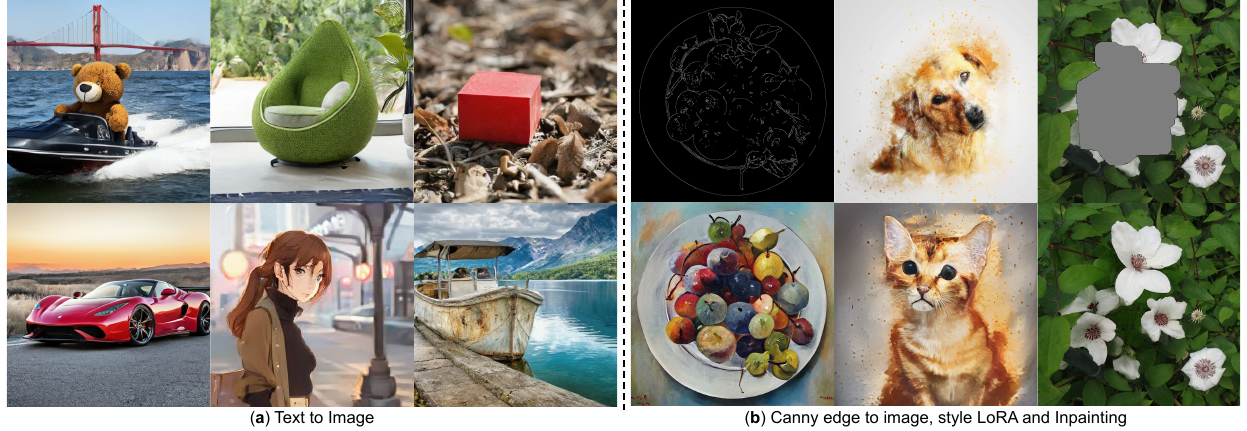}
    \caption{MobileDiffusion for (a) Text to image generation. (b) Canny edge to image, style LoRA and inpainting. Samples are all generated in one-step.}
    \label{fig:teaser}
\end{figure}
\begin{abstract}
The deployment of large-scale text-to-image diffusion models on mobile devices is impeded by their substantial model size and high latency. In this paper, we present \textbf{MobileDiffusion}, an ultra-efficient text-to-image diffusion model obtained through extensive optimizations in both architecture and sampling techniques. We conduct a comprehensive examination of model architecture design to minimize model size and FLOPs, while preserving image generation quality. Additionally, we revisit the advanced sampling technique by diffusion-GAN, and make one-step sampling compatible to downstream applications trained on the base model. Empirical studies, conducted both quantitatively and qualitatively, demonstrate the effectiveness of our proposed technologies. With them, MobileDiffusion achieves instant text-to-image generation on mobile devices, establishing a new state of the art.
\end{abstract}

\section{Introduction} \label{sec:intro}
Text-to-image diffusion models \cite{nichol2022glide,ramesh2022hierarchical,saharia2022photorealistic,balaji2022ediffi,rombach2022high,podell2023sdxl} have exceptional capabilities in generating high-quality images conditioned on texts. These models serve as the foundation for a variety of applications, including image editing\cite{hertz2022prompt,tumanyan2023plug}, controlled generation \cite{zhang2023adding, mou2023t2i}, personalized content generation~\cite{ruiz2023dreambooth,gal2022image}, video synthesis\cite{blattmann2023align,guo2023animatediff}, and low-level vision tasks\cite{xu2023open, li2023your}. These large-scale models are essentially considered for necessarily running on servers with powerful neural compute units. Only a few work~\cite{li2023snapfusion,kim2023architectural} has barely touched running diffusion models on mobile devices, which remains an open challenge.

We identify two primary factors causing the inefficiency of text-to-image diffusion models. Firstly, the complexity of the network architecture in text-to-image diffusion models involves a substantial number of parameters, often reaching to the billions, resulting in computationally expensive evaluations. Secondly, the inherent design of diffusion models requires iterative denoising to generate images, necessitating multiple evaluations of the model \cite{ho2020denoising, song2020score}. These inefficiency challenges pose a significant barrier to deploy in resource-constrained environments, such as on mobile devices. As a result, despite the potential benefits, such as enhancing user experience, addressing emerging privacy concerns, and saving cost, this aspect remains relatively unexplored within the current literature.

In this paper, we aim to develop an ultra-efficient diffusion model suitable for a range of on-device applications, necessitating a design that is lightweight, rapid, and versatile enough to handle various downstream generative tasks, including in-painting, controllable generation, and personalized generation efficiently. We approach the identified challenges—architectural and sampling efficiency—through a divide-and-conquer strategy, addressing each separately.

The challenge of \textbf{architectural efficiency} in diffusion models has been rarely addressed in existing literature. Prior attempts have focused on eliminating redundant neural network blocks \cite{li2023snapfusion,kim2023architectural} or reorganizing them to improve efficiency \cite{hoogeboom2023simple} but lack a comprehensive analysis of the model architecture's components. Our research fills this gap by conducting an in-depth review of diffusion networks, leading to a carefully optimized model architecture. With fewer than 400 million parameters, our model achieves high-quality image generation, surpassing previous efforts in efficiency.

The second challenge, namely the \textbf{sampling efficiency} of diffusion models, has been an active research area. With advanced numerical solvers\cite{lu2022dpm,lu2023dpm,bao2022analytic,karras2022elucidating,song2020denoising} or distillation techniques \cite{salimans2021progressive,meng2023distillation,song2023consistency,li2023snapfusion,luo2023latent}, the necessary number of network evaluations of sampling has been reduced significantly from several hundreds to less than 10. Despite these advances, the reduced evaluation steps can still pose challenges on mobile devices, where even minimal processing demands may prove too burdensome. Our attention, therefore, turns towards pioneering single-step diffusion generation methods \cite{ufogen2023, sauer2023adversarial, yin2023one}. These advancements have the potential to transform diffusion models into efficient one-step text-to-image generators. Yet, there remains an unmet need for these models to flexibly accommodate a variety of conditional generation tasks through the integration of additional modules trained separately. To close this gap, we explore the nuanced design space of diffusion-GAN hybrids \cite{ufogen2023,sauer2023adversarial,lin2024sdxl,xiao2021tackling}, a forefront category of one-step diffusion models gaining popularity. Specifically, our examination centers on the UFOGen framework \cite{ufogen2023}, a recent innovation in this domain. Our work advances the discourse by conducting an in-depth analysis of UFOGen's objective function and training methodologies, culminating in a refined approach for developing highly effective one-step diffusion GAN models. This endeavor aims to achieve a dual objective: to facilitate the creation of high-quality text-to-image synthesis and to ensure the seamless adaptability of the model for diverse downstream applications, thereby marking a significant stride towards realizing ultra-fast, versatile one-step diffusion models.

Our combined efforts in architectural design and sampling efficiency enable instant generation of high-quality $512\times512$ images on mobile devices, notably achieving \textbf{0.2 second} on an iPhone 15 Pro, which is about the average response time of human to visual stimulus. This achievement marks a significant leap over previous state-of-the-art in on-device text-to-image generation \cite{li2023snapfusion}. We introduce our model as MobileDiffusion, highlighting its potential as a foundational generative model for edge devices.

Our paper makes several key contributions to the field of on-device generative modeling, detailed as follows:
\begin{enumerate}
    \item We present a comprehensive exploration of architectural efficiency in text-to-image diffusion models. Our work introduces a refined diffusion model architecture that is not only highly efficient and lightweight but also demonstrates rapid performance on mobile devices. 
    \item  We study the design space of one-step diffusion-GAN models to present the best recipe for training such models, leading to high generative capabilities beyond text-to-image applications in one-single step. To our knowledge, our work pioneers the adaptation of a one-step diffusion model for a broad range of applications.
    \item Synthesizing our efforts, we introduce \textit{MobileDiffusion}, an ultra-efficient diffusion model framework specifically designed for on-device deployment. 

\end{enumerate}
\section{Related works} \label{sec:related_works}
As elaborated in Section \ref{sec:intro}, to improve the inference efficiency of text-to-image diffusion models and ultimately enable their deployment on mobile devices, there are two primary areas of focus: \textit{architecture efficiency} and \textit{sampling efficiency}. We briefly review the prior work.

\textbf{Architecture Efficiency }
A limited number of prior works have discussed the architectural efficiency of diffusion models. Early diffusion models~\cite{dhariwal2021diffusion,rombach2022high,saharia2022photorealistic} adopted a UNet structure with transformer blocks. Recent works~\cite{bao2023all,peebles2023scalable} started to introduce vision transformers to diffusion. \cite{hoogeboom2023simple} proposed UViT, a UNet structure with a transformer backbone. With transformers establishing their advances, the differences of these methods lie in approaches to balance computation and cost, through UNet or patchify.
 \cite{kim2023architectural} introduced an approach to distill larger diffusion models into smaller student models by selectively removing specific blocks from the teacher model. Meanwhile, \cite{li2023snapfusion} presented an efficient architecture search method. They trained a UNet with redundant blocks using robust training techniques \cite{huang2016deep, yu2018slimmable} and then pruned certain blocks based on metrics, resulting in an architecture suitable for distillation. 
In our work, we align with the insights of \cite{hoogeboom2023simple} regarding the transformer components in the UNet while introducing distinctive perspectives elaborated in Section \ref{sec:method_md}.
In contrast to approaches like \cite{li2023snapfusion,kim2023architectural}, our focus extends beyond mere block removal. Instead, we conduct a more nuanced analysis and modification of the architecture. Additionally, unlike \cite{li2023snapfusion,kim2023architectural}, we opt not to employ knowledge distillation from a larger model. Our observations indicate that training our model from scratch yields satisfactory results.

\textbf{Sampling Efficiency}
Diverse strategies have emerged to reduce the required sampling steps of diffusion models, broadly falling into two categories. The first involves developing fast solvers to more efficiently solve the differential equation tied to the denoising process, thereby reducing the necessary discretization steps  steps~\cite{lu2022dpm,lu2022dpm,karras2022elucidating,bao2022analytic,dockhorn2022genie}. The second approach leverages knowledge distillation techniques to compress the sampling trajectory ~\cite{meng2021sdedit,salimans2021progressive,song2023consistency,li2023snapfusion,luo2023latent}. Among them, consistency distillation \cite{song2023consistency,luo2023latent,luo2023lcm} have shown promise in reducing the number of sampling steps to 4-8, while still facilitating downstream generative tasks, as highlighted by \cite{xiao2023ccm}. Nonetheless, attempts to further decrease the number of steps often result in diminished output quality. Recent innovations aim to revolutionize this area by enabling single-step generation \cite{liu2023instaflow,ufogen2023, sauer2023adversarial, yin2023one}. One important direction of research in one-step diffusion is diffusion-GAN hybrids \cite{xiao2021tackling}, namely fine-tuning diffusion models with an adversarial objective \cite{ufogen2023, sauer2023adversarial}. Despite these advances, the adaptation of one-step diffusion models for a broader array of downstream tasks remains an under-explored territory. A concurrent work \cite{lin2024sdxl} employs parameter-efficient adversarial fine-tuning. However, their resulting parameter-efficiently fine-tuned model does not facilitate one-step generation.

\section{Designing a mobile-friendly diffusion architecture}\label{sec:method_md}
In this section, we present our recipe for crafting highly efficient text-to-image diffusion models, which ultimately lead to sub-second generation on mobile devices. Following~\cite{rombach2022high,podell2023sdxl}, we adopt latent diffusion for its efficiency of learning text-guided generation in latent space. Our model design lies in an efficient diffusion network and a lightweight VAE decoder. We employ the CLIP-ViT/L14~\cite{radford2021learning} as the text encoder, which can run in a couple of milliseconds on-device.



\begin{figure*}[!t]
    \centering
        \includegraphics[width=0.38\textwidth]{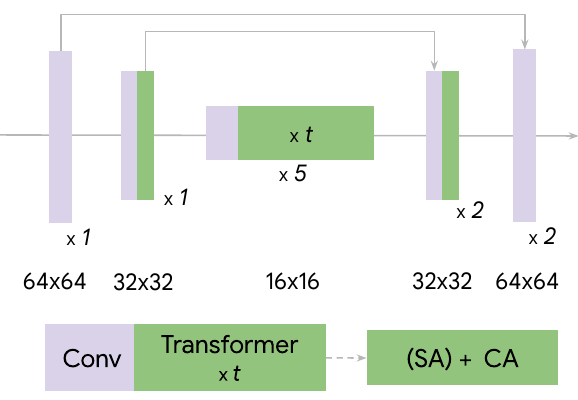} \hspace{20pt}
        \includegraphics[width=0.32\textwidth]{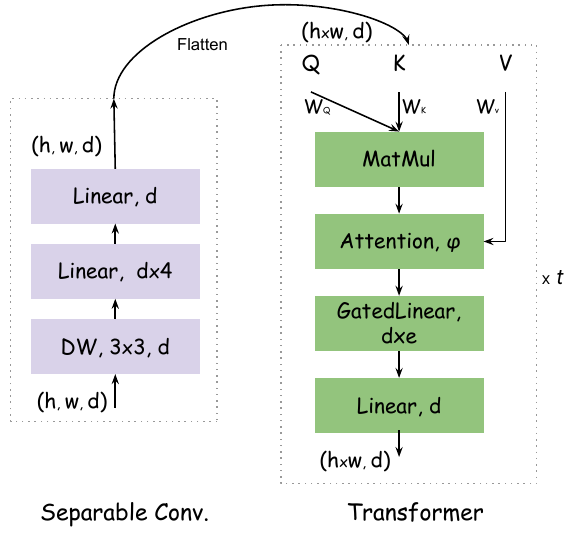} \\
    \caption{Model illustration of MobileDiffusion. ($\mathsf{Conv}$: Convolution. $\mathsf{SA}$: Self-attention (optional). $\mathsf{CA}$: Cross-attention. $\phi$: Non-linearity to calculate attention weights. $e$: Expansion factor in gated linear layer~\cite{shazeer2020glu}. $\mathsf{DW}$: Depth-wise convolution.)}
    \label{fig:unet}
\end{figure*}

\subsection{Diffusion network}
Early diffusion models~\cite{dhariwal2021diffusion,rombach2022high,saharia2022photorealistic} adopted a UNet architecture with convolutions and attentions. Recent works~\cite{bao2023all,peebles2023scalable} introduced vision transformer (ViT) backbones to diffusion with a great potential in scaling up the models. We noticed that transformers are expensive for a large sequence length. It needs to be considered when designing models with budget constraints. SnapFusion~\cite{li2023snapfusion} removes transformer blocks at the highest resolution in latent space. ViT~\cite{peebles2023scalable} patchifies a spatial input to a sequence of tokens, with a reduced length. 

We follow the UViT~\cite{hoogeboom2023simple} for designing our diffusion network with text guidance. As shown in Fig.~\ref{fig:unet}, we leverage the UNet to reduce the sequence length and compute transformers efficiently. We made novel changes to the UViT architecture by conducting a comprehensive investigation of two fundamental building blocks: transformer and convolution. Throughout the study, we control the training pipeline (\eg data, optimizer) to study the effects of different architectures. We optimize the network towards lower numbers of parameters and floating point operations (FLOPs), while preserving the quality. We leverage two metrics, Fréchet Inception Distance (FID) and CLIP-ViT B/32 on MS-COCO 2014 30K, complemented by visual inspections to monitor the quality of each model variation.

\subsection{Optimize transformer}

\vspace{3mm} \noindent \textbf{Scale up the backbone.} 
Model optimization does not always indicate scaling down. By relocating model parameters, we can achieve more efficient compute. We noticed that transformers are efficient at low resolutions. Therefore, we scale up the transformer backbone in the UViT by moving more transformer layers to the bottleneck, while maintaining the total number of parameters. Additionally, the computation complexity of self-attention is $O(nd^2+n^2d)$, where $n$ is the sequence length, and $d$ is the channel dimension. At low resolutions (\eg $16\times16$), the channel dimension contributes more to the computation. Our empirical observations indicate that slightly reducing the channel dimension in the bottleneck does not adversely impact the quantitative metrics or the visual quality of the generated samples. On the other hand, attempting to stack more transformer blocks with a large reduction of channel dimension proves detrimental, resulting in an evident degradation in visual quality characterized by poor object composition and intricate artifacts. We found $1024$ is the best channel dimension with a good trade-off of efficiency and quality.

\vspace{3mm} \noindent \textbf{Decouple SA from CA.} 
In text-to-image diffusion models, self-attention plays a pivotal role in capturing long-range dependencies, although it comes with significant computational costs at higher resolutions. For instance, at a resolution of $32\times32$, the sequence length for self-attention is $1024$. In prior work~\cite{hoogeboom2023simple,podell2023sdxl}, self-attention and cross-attention layers are moved together when approaching efficient designs. Upon further investigation, we discover that retaining cross-attention layers while discarding only the self-attention layers at high resolutions does not result in a performance drop. We conjecture that cross-attention is useful across different resolutions, as text guidance is crucial for both the global layout and local texture of the image. Notably, the computational cost of cross-attention at high resolution is significantly lower than that of self-attention, due to the smaller sequence length of text embeddings (\eg 77 for SD). Consequently, removing only the self-attention layer yields a substantial efficiency boost. In light of these insights, we adopt a design that maintains performance while enhancing efficiency: entirely removing transformer blocks at the highest resolution ($64\times64$); eliminating self-attention layers in the transformer blocks at $32\times32$ resolution and the outer $16\times16$ stack; retaining complete transformer blocks in the inner $16\times16$ stack and the innermost bottleneck stack. 

\vspace{3mm} \noindent \textbf{Share key-value projections.}
Within an attention layer, both the key and value are projected from the same input $x$, denoted as $K = x \cdot W_K$ and $V = x \cdot W_V$. Our experiments reveal that adopting a parameter-sharing scheme, specifically setting $W_K = W_V$ for self-attention layers, does not adversely affect model performance. Therefore, we choose to implement this parameter-sharing strategy, resulting in approximately a $5\%$ reduction in parameter count.

\vspace{3mm} \noindent \textbf{Replace $\mathsf{gelu}$ with $\mathsf{swish}$.} The GLU (Gated Linear Unit)~\cite{shazeer2020glu} adopts the $\mathsf{gelu}$ activation function, a choice that unfortunately introduces numerical instability issues when employed in float16 or int8 inference on mobile devices due to the involvement of a cubic operation in the approximation~\cite{choi2023squeezing}. Additionally, the $\mathsf{gelu}$ activation incurs slower computation, relying on specific hardware optimizations~\cite{wang2022towards,sun2020mobilebert}. Therefore, we propose substituting $\mathsf{gelu}$ with $\mathsf{swish}$, which maintains a similar shape but is more cost-effective and computationally efficient. Importantly, empirical results indicate that this replacement does not lead to a degradation in metrics or perceptual quality.

\vspace{3mm} \noindent \textbf{Finetune $\mathsf{softmax}$ into $\mathsf{relu}$.} 
Given key $K$, query $Q$ and value $V$, attention calculates $\mathbf{x} = \phi(K^\mathsf{T} Q) V$, where the function $\phi(\cdot)$ is the $\mathsf{softmax}$. However, the $\mathsf{softmax}$ function, where $\mathsf{softmax}(\mathbf{x}) = e^{\mathbf{x}} / \sum^N_{j=1}e^{x_j}$ and $\mathbf{x}=(x_0, x_1, ..., x_N)$, is computationally expensive due to non-efficient parallelization of exponentiation and summation across the sequence length. Conversely, point-wise activations like $\mathsf{relu}$ present a quicker alternative that does not rely on specific hardware optimizations, and it can serve as a viable substitute~\cite{wortsman2023replacing}. Consequently, we propose employing $\mathsf{relu}$ in our attention computation. An intriguing finding is that there is no need to train a $\mathsf{relu}$-attention model from scratch; instead, finetuning from a pre-trained $\mathsf{softmax}$-attention model proves sufficient, and this fine-tuning process can be accomplished quickly, for example, within 10,000 iterations. Visual ablation results for this modification are provided in Appendix.

\vspace{3mm} \noindent \textbf{Trim feed-forward layers.} 
In feed-forward layers of a transformer, the expansion ratio is default to 4, which is further doubled with gated units~\cite{shazeer2020glu}. This amplifies the parameter count significantly. For example, a channel dimension 1280 surges to 10240 after projection. Such high dimensionality can be limiting in resource-constrained mobile applications. After thorough ablations, we found that trimming the expansion ratio to 3 results in nearly identical performance. With this adjustment, the FID score experiences only a slight increase of 0.27, while concurrently leading to $10\%$ reduction in parameters count.

\subsection{Optimize convolution}
\vspace{3mm} \noindent \textbf{Separable convolution.} Residual blocks in vanilla convolutional layers typically involve a substantial number of parameters, prompting various endeavors to enhance the parameter efficiency of convolutional layers. One proven approach in this context is separable convolution \cite{howard2017mobilenets}.
We have observed that replacing vanilla convolution layers with lightweight separable convolution layers in the deeper segments of the UNet yields similar performance. As a result, we replace all the convolutional layers in the UNet with separable convolutions, except for the outermost level.  The separable convolutional block we adopted, as illustrated in Figure~\ref{fig:unet}, shares similarities with ConvNeXt~\cite{liu2022convnet} but employs a smaller $3\times3$ kernel size. While we experimented with larger kernel sizes, such as $7\times7$ and $9\times9$, we found that they did not provide additional improvements.

\vspace{3mm} \noindent \textbf{Prune redundant residual blocks.} Convolution operations on high-resolution feature maps are especially computationally expensive, and pruning is a straightforward way to improve model efficiency. Through a comprehensive network search, we reduce the number of required residual blocks from 22 (in SD) to more efficient and streamlined 12. More specifically, we set 1 layer per block instead of 2 in SD, except for the innermost blocks. While maintaining a balance between resource consumption and model performance.

\setlength{\tabcolsep}{6pt} 
\begin{table*}[!t]
    \centering
    \scalebox{0.8}{
    \begin{tabular}{l|ccccc}
    \toprule
        Models & \#Channels & \#ConvBlocks & \#(SA+CA) & $\#$Params(M) & $\#$GFLOPs \\\midrule
        SD-XL~\cite{podell2023sdxl}  & (320, 640, 1280) & 17 & 31+31 & 2,600 & 710 \\
        SD-1.4/1.5~\cite{rombach2022high}  & (320, 640, 1280, 1280) & 22 & 16+16 & 862 & 392 \\
        SnapFusion~\cite{li2023snapfusion}   & (320, 640, 1280, 1280) & 18 & 14+14 & 848 & 285 \\
        DiT XL/2~\cite{peebles2023scalable} & (1152) & 0 & 28+0 & 675 & 525\\\midrule
        MobileDiffusion  & (320, 640, 1024) & 11 & 15+18 & 386  & 182\\
        \bottomrule
    \end{tabular}
    }
    \caption{Comparison with other recognized latent diffusion models.}
    \label{tab:tigo_variations}
\end{table*}

\subsection{Model details}
Employing the optimization techniques discussed earlier, we meticulously refine our selection of viable architecture candidates by imposing upper bounds on the numbers of parameters and FLOPs. We aim at 400 million parameters and 200 GFLOPs to achieve instant generation on devices. Notably, for the ease of model search, we combine the innermost level of the down stack, the up stack, and the middle bottleneck into a unified module, as they share the same feature map dimension. This corresponds to the 5 layers in the middle of architecture as illustrated in Figure~\ref{fig:unet}. Within this constrained search space, we systematically explore variations in the number of transformer layers and innermost channel dimensions, concentrating on the pivotal architectural parameters that significantly influence model performance. Due to the computational demands associated with training diffusion models, we prioritize optimizing these critical parameters to enhance the model's efficiency and effectiveness. In Table~\ref{tab:tigo_variations}, we present a comparative analysis of MobileDiffusion, against SD UNets, SnapFusion UNet~\cite{li2023snapfusion} optimized for on-device, and DiT XL/2~\cite{peebles2023scalable}. It underscores that MobileDiffusion exhibits the highest efficiency. MD conveys more computation from convolution to attention than UNets, and embodies fewer FLOPs than pure transformer architectures. 
\vspace{-8mm}
\setlength{\tabcolsep}{0pt} 
\renewcommand{\arraystretch}{1} 
\begin{figure}[!h]
    \centering
    \scalebox{0.85}{
    \begin{tabular}{cccc}
        \includegraphics[width=0.25\linewidth,trim={4cm 9cm 3cm 0},clip]{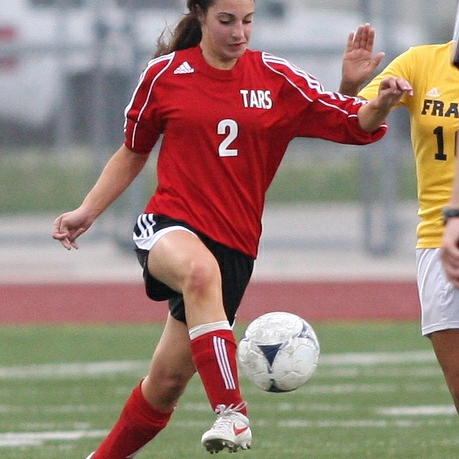} &
        \includegraphics[width=0.25\linewidth,trim={4cm 9cm 3cm 0},clip]{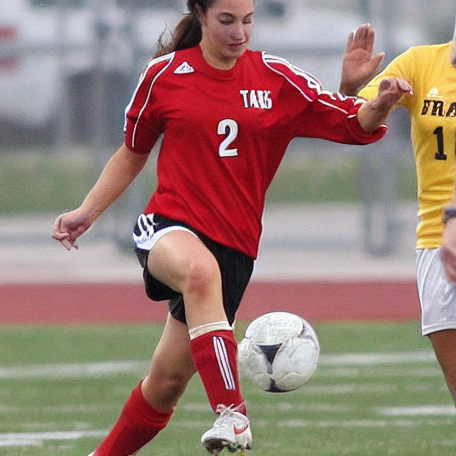} &
         \includegraphics[width=0.25\linewidth,trim={4cm 9cm 3cm 0},clip]{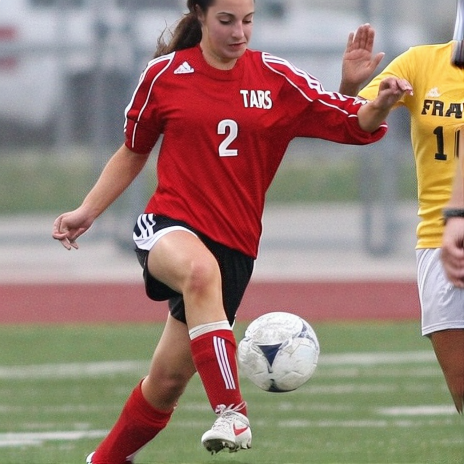} &
        \includegraphics[width=0.25\linewidth,trim={4cm 9cm 3cm 0},clip]{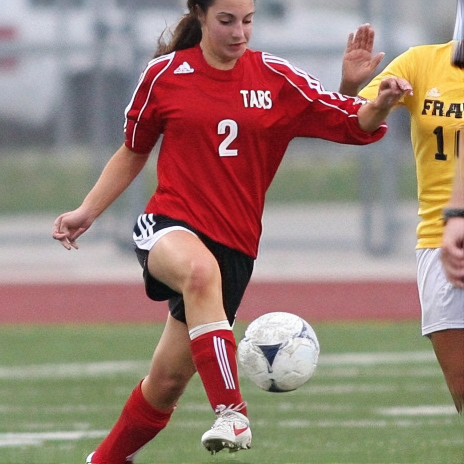} \\
        \footnotesize{Input} & \footnotesize{SD} & \footnotesize{Ours} & \footnotesize{Ours-distilled} \\
    \end{tabular}
    }
    \caption{VAE reconstruction comparison among SD decoder, our decoder and our distilled decoder.}
    \label{fig:vae_compare}
\end{figure}
\vspace{-6mm}
\subsection{Optimize VAE}

In latent diffusion, VAE converts a $H\times W\times3$ image to a $\frac{H}{f}\times\frac{W}{f}\times c$ latent. SD~\cite{rombach2022high} uses $f=8$ and $c=4$, while EMU~\cite{dai2023emu} uses $f=8$ and $c=16$. Using smaller channel number $c$ results in better compression but worse reconstruction. After an empirical study, we choose $f=8$ and $c=8$ for training our VAE. This modification enhances the quality of image reconstruction, as demonstrated in Figure~\ref{fig:vae_compare}. Similar to other VAEs used for latent diffusion models, our VAE is trained with a combination of loss functions, including $\mathcal{L}_2$ reconstruction loss, KL divergence for regularization, perceptual loss, and adversarial loss. We train the VAE with batch size 256 for 2 million iterations using the same dataset for diffusion model training. As pointed out in~\cite{li2023snapfusion}, when evaluating with a smaller number of diffusion steps, image decoder becomes a non-negligible component. To further enhance efficiency, we design a lightweight decoder architecture by pruning model \textit{width} and \textit{depth}. We train the lightweight decoder for 400K iterations with the encoder frozen. We remove the unnecessary KL regularization in the objective, and reduce the weight of adversarial loss. The distilled decoder leads to a significant performance boost, 3 times faster than SD's decoder as shown in Table~\ref{tab:model_parts}.

\section{Elucidating the design space of UFOGen}\label{sec:method_ufo}
In this section, we detail our studies on the ultra-fast one-step diffusion models, focusing on enhancing image quality and, crucially, enabling the integration of a broad spectrum of downstream applications into the one-step generation paradigm. Our investigation primarily concentrates on UFOGen \cite{ufogen2023}, an innovative adversarial fine-tuning approach that has shown considerable promise for one-step text-to-image generation. We start with a concise overview of UFOGen, followed by a discussion on the series of ablation studies we undertook to refine its training methodology

\subsection{Overview of UFOGen}
UFOGen employs adversarial fine-tuning of a pre-existing diffusion model to achieve one-step generation. This process involves the one-step generator $G_{\theta}$, being trained jointly with a discriminator $D_{\phi}$ with the following objective 
\begin{align}\label{eq:ufogen}
        &\min_\theta\max_{\phi}\mathbb{E}_{q(x_0)q(x_{t-1}|x_0), p_{\theta}(x_{0}')p_{\theta}(x_{t-1}'|x_0')} \Bigl[\\ \nonumber
    &\underbrace{[\log(D_{\phi}(x_{t-1}, t))]+ [\log(1-D_{\phi}(x'_{t-1}, t))]}_{\text{adversarial loss}}
    + \underbrace{\lambda \gamma_t\left\lVert x_0-x_0'\right\rVert^2}_{\text{diffusion loss}} \Bigl],
\end{align}
where $q(x_0)$ is the training data distribution, $q(x_{t-1}|x_0)$ is the forward diffusion process. The generator is parameterized by generating a clean image $x_0' = G_{\theta}(x_t, t) \sim p_{\theta}(x_0')$, and $x_{t-1}' \sim p_{\theta}(x_{t-1}'|x_0')$ is applying forward diffusion process on generator's output $x_0'$. Besides the adversarial loss, UFOGen also includes the original diffusion loss in its objective, and this is shown to be crucial for stabilizing the training. The generator and discriminator both have the same structure and initialized by the pre-trained diffusion model. The UNet discriminator aggregates logits across output locations make decisions. 

UFOGen has demonstrated its efficacy in generating high-quality images from text prompts in a single step, significantly enhancing the efficiency of sampling in diffusion models. However, training strategies of UFOGen have not been well-studied. In addition, seamlessly incorporating the separately-trained modules such as LoRA \cite{hu2021lora}, ControlNet \cite{zhang2023adding} and T2I-adapter \cite{mou2023t2i} with one-step sampling remains unexplored. While the original UFOGen study \cite{ufogen2023} showcases successful image generation from depth maps and canny edges, these outcomes are achieved through task-specific adversarial training, necessitating a separate copy for each new conditional task — far from the ideal scenario. To this end, we carefully studied the design space of UFOGen, drawing motivations from related methods. Our finalized design choice is obtained by comparing different variants on both the text-to-image generation and downstream conditional generation tasks.  

\subsection{Reconstruction term in the training objective}
The first design aspect worth exploring is the role of the reconstruction term in UFOGen's training objective. The formulation of UFOGen, as specified in Equation \ref{eq:ufogen}, integrates both adversarial and reconstruction losses. The purpose of the adversarial loss is to enhance the visual accuracy of images produced through a singular diffusion step. Traditionally, a diffusion model's single-step generation process is calibrated to forecast the \textit{expected} pristine image from a given noisy input, symbolized by $\mathbb{E}[x_0|x_t]$. As a result, this often yields images that are somewhat blurry, lacking in fine details due to the inherent averaging across the data distribution. The inclusion of an adversarial loss term empowers the model to generate sharper images, particularly at elevated noise levels. Conversely, the functionality of the reconstruction loss within the training regime is more nuanced. It is postulated to act as a stabilizing force during the adversarial training phase, offering a straightforward learning directive. Beyond this, we propose that it plays a vital role in maintaining the integrity of the pre-trained diffusion model's characteristics. Such preservation is crucial; in its absence, the one-step model risks deviating markedly from the original feature space of the pre-trained model. Such a divergence could severely limit the compatibility with downstream modules specifically tailored for the original diffusion framework.

Our conjecture suggests it might be helpful to provide enhanced regularization to encourage the preservation the pre-trained model's characteristics, ensuring that the one-step model remains closely aligned with the foundational diffusion model's feature landscape. To this end, we examined some alternative options to the reconstruction loss in Equation \ref{eq:ufogen}. The first one is a distillation loss that explicitly align the features of UFOGen's generator to be aligned with the teacher model:
\begin{align}\label{eq:distill_loss}
    \mathcal{L}_{\text{distill}} = \left\lVert \text{sg}\left(G^{\text{teacher}}(x_t, t)\right) - x_0'\right\rVert^2,
\end{align}
where $G^{\text{teacher}}_{\theta}$ is the pre-trained teacher model, and $\text{sg}(\cdot)$ stands for stop gradient. Note that the distill loss is similar to the regularization term in the objective of Adversarial Diffusion Distillation (ADD) \cite{sauer2023adversarial}. 

More generally, we also test the EMA distillation objective, which is formalized as follows
\begin{align}\label{eq:ema_loss}
    \mathcal{L}_{\text{ema}} = \left\lVert \text{sg}\left(G^{\text{EMA}}_{\theta}(x_t, t)\right) - x_0'\right\rVert^2,
\end{align}
where $G^{\text{EMA}}_{\theta}$ is the exponential moving average of $G_{\theta}$. Given that $G_{\theta}$ originates from a pre-trained model, the EMA mechanism effectively retains a substantial portion of the pre-trained model's information. This loss can also ensure that the essence of the pre-trained model is conserved, while providing more flexibility. Note that the distillation loss in Equation \ref{eq:distill_loss} is a special case of our loss, with the EMA decay parameter being set to 1. We conduct comprehensive ablation studies on these loss terms in Section \ref{sec:exp_ablate}.

\subsection{Parameter-efficient adversarial fine-tuning}
Current methods in adversarial diffusion fine-tuning, exemplified by UFOGen \cite{ufogen2023} and ADD \cite{sauer2023adversarial}, adjust the entire model. This extensive parameter modification can lead to significant shifts in the internal feature representation, potentially undermining the model's ability to work effectively with downstream tasks. Parameter-efficient fine-tuning emerges as a compelling solution, drawing from its successful application across both the visual and language fields \cite{hu2021lora,ding2022delta}. Among these methods, LoRA (Low-Rank Adaptation) fine-tuning \cite{hu2021lora} stands out for its ability to refine diffusion models \cite{luo2023lcm}, offering a balanced approach that supports rapid sampling while maintaining task adaptability. 

Motivated by the success of LCM-LoRA, we explore the potential of integrating LoRA fine-tuning within the UFOGen training framework. We apply LoRA exclusively to the generator, initializing the model weights from the pre-trained diffusion framework while setting the LoRA layers to initialize with a null effect. The discriminator is still initialized from the pre-trained diffusion model.

\section{Experiments} \label{sec:experiments}

We begin by providing an overview of the training specifics in Section~\ref{sec:training}, which encompass experimental setup, dataset, and the evaluation protocol. In Section~\ref{sec:main_results}, we present the primary results of our text-to-image generation including quantitative metrics, generated samples, perceptual assessments, as well as on-device benchmarks. In Section~\ref{sec:application}, we delve into an exploration of various mobile applications, such as the integration of controls through plugins and of personalization capabilities via LoRA finetuning.


\subsection{Training Details}\label{sec:training}

\vspace{3mm} \noindent \textbf{Dataset}
Our training process leverages a proprietary dataset, comprising an extensive collection of 150 million image-text pairs from public web~\cite{chen2023pali}. The majority of images have a resolution greater than $256\times256$, and 40 million images with a resolution over $512\times512$. To prepare these images for training, we follow a consistent preprocessing methodology.

\vspace{3mm} \noindent \textbf{Optimization}
Our models are trained with the AdamW optimizer~\cite{loshchilov2018decoupled}, configured with a learning rate of 0.001, $\beta_1$ of 0.9, $\beta_2$ of 0.999, and a weight decay of 0.01. We adopt a progressive training strategy: first, we train at $256\times256$ resolution with batch size 4096 for 0.75M steps to capture high-level semantics, followed by an additional 0.25M steps at a resolution of $512\times512$ with batch size 2048 to refine features.

\vspace{2mm} \noindent \textbf{Training Cost}
When searching and selecting superior candidates, we rely on the FID and CLIP scores reported at 30K training steps to decide the candidates to proceed. This corresponds to approximately 4 hours of computation using 32 TPUs with 16GB memory. We have designed a stopping mechanism for our experiments, which halts when the FID and CLIP scores show signs of slower improvement compared to previous candidates. This strategy works reasonable well in the process.
In total, our endeavor consumes approximately 512 TPUs spanning 15 days to complete the network search.

\vspace{3mm} \noindent \textbf{Metrics}
We employed the MS-COCO dataset~\cite{lin2014microsoft} as the primary source for our evaluations. In line with established practices, we present results for the zero-shot FID-30K scenario, where 30,000 captions are randomly selected from the COCO validation set. These captions serve as inputs for the image synthesis process. We calculate the FID score~\cite{heusel2017gans} to gauge the dissimilarity between the generated samples and the 30,000 reference ground truth images. Additionally, we provide the CLIP score~\cite{radford2021learning}, which assesses the average similarity between the generated samples and their corresponding input captions by utilizing features extracted from a pre-trained CLIP model, OpenCLIP-ViT/g14~\cite{cherti2023reproducible}.

\subsection{Text-to-Image Generation}\label{sec:main_results}
We present comprehensive results of the efficiency and quality of text-to-image generation, spanning quantitative and qualitative comparisons, and on-device benchmarks.

\begin{wrapfigure}[8]{C}{0.32\textwidth}
\vspace{-10mm}
\begin{minipage}{0.3\textwidth}
    \centering
    \begin{tabular}{c}
        \includegraphics[width=0.95\textwidth]{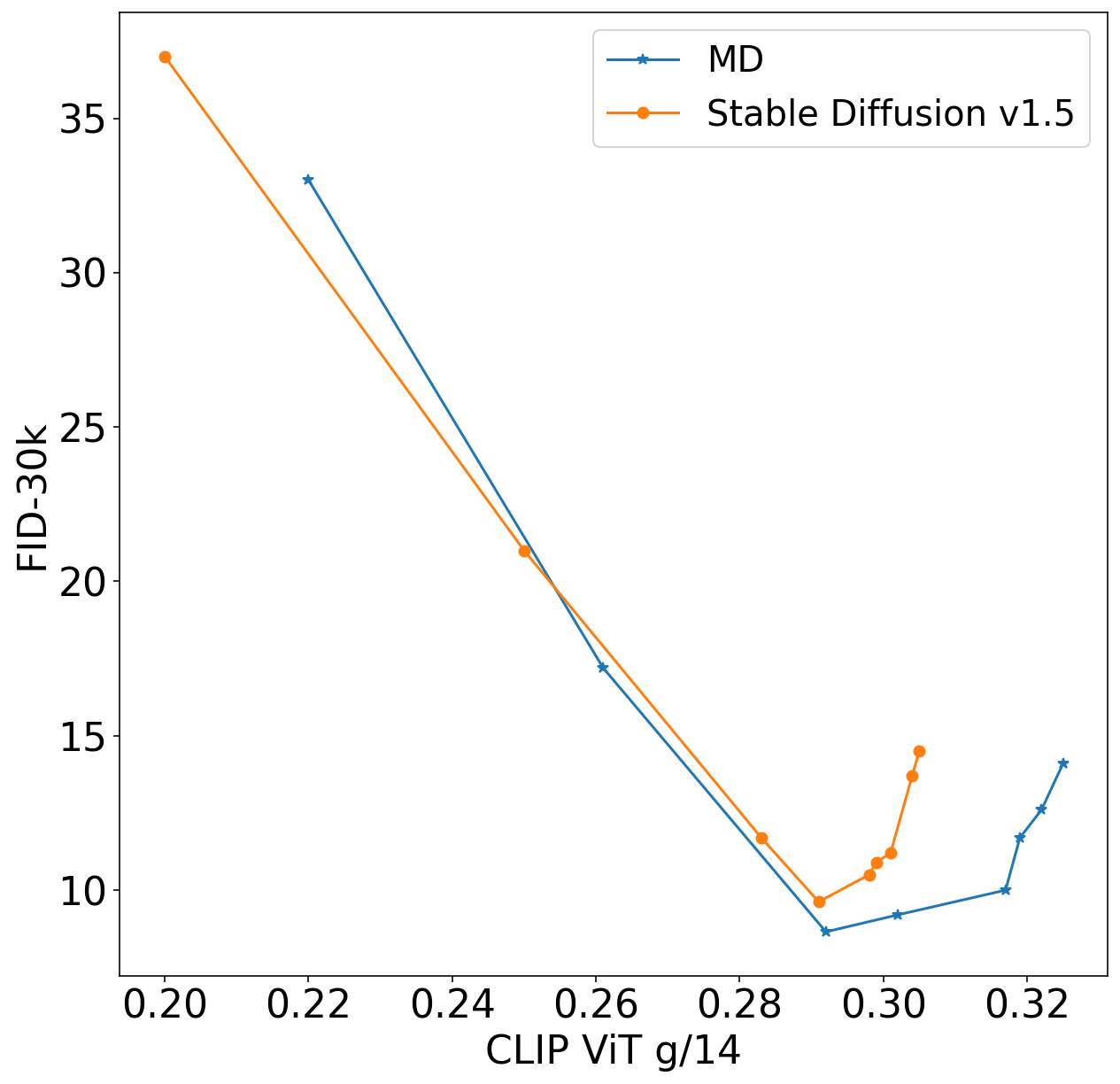}
    \end{tabular}
    \vspace{2mm}
\end{minipage}
\end{wrapfigure}
\vspace{3mm} \noindent \textbf{Quantitative Evaluation} In Table~\ref{tab:quantitative_cmp}, we conducted a comprehensive comparison of various text-to-image generation models. For DDIM, we achieved the lowest FID scores by adjusting the \textit{cfg} scales around 3.0. Since UFOGen does not have \textit{cfg}, we report the FID scores as is. We also compare CLIP scores of models close to ours. While being compact in model size, our MobileDiffusion achieves better metrics than previous solutions of SnapFusion~\cite{li2023snapfusion} and UFOGen~\cite{ufogen2023}. The figure on the right shows the FID vs. CLIP curve by comparing MobileDiffusion and Stable Diffusion v1.5 using 50-step DDIM by varying cfg. 

We also consider human preference evaluation using the standard HPS v2 benchmark~\cite{wu2023human}. We compare with SD v2.0 and v1.4 in Table~\ref{tab:hpsv2}. Despite with much smaller model size and number of inference steps, our model obtained comparable performance with Stable Diffusion variants. 

\setlength{\tabcolsep}{6pt} 
\begin{table}[!t]
    \centering
    \scalebox{0.75}{
    \begin{tabular}{llrcccc}
    \toprule
        Models & Sampling &  \#Steps & FID-30K$\downarrow$ & CLIP$\uparrow$ & \#Params(B) & \#Data(B) \\ \midrule
        GigaGAN~\cite{kang2023scaling} & 1-step & 1 & 9.09 & - & 0.9 &  0.98 \\
        LAFITE~\cite{Zhou_2022_CVPR} & 1-step &  1 & 26.94 & -  & 0.23 & 0.003 \\\midrule         
        DALL$\cdot$E-2~\cite{ramesh2022hierarchical} & DDPM & 292 & 10.39 & -  & 5.20 & 0.25 \\ 
        Imagen~\cite{saharia2022photorealistic} & DDPM &  256 & 7.27 & -  & 3.60 & 0.45 \\
        SD~\cite{rombach2022high} & DDIM & 50 & 9.62 & 0.304 & 0.86 & 0.60 \\
        PIXART-$\alpha$~\cite{chen2023pixart} & DPM&  20 & 10.65 & - & 0.6 & 0.025 \\
        BK-SDM~\cite{kim2023bksdm} & DDIM & 50 & 16.54 & -  & 0.50 & -\\ \midrule
        SnapFusion~\cite{li2023snapfusion} & Distilled & 8 & 13.5 & 0.308 & 0.85 & -\\ 
        UFOGen~\cite{ufogen2023} & 1-step & 1& 12.78 & 0.317 & 0.86 & 0.6 \\ \midrule
        \multirow{2}{*}{\textbf{MoibleDiffusion}} & DDIM & 50 & 8.65 & 0.325 & \multirow{2}{*}{0.39} & \multirow{2}{*}{0.15} \\
        & 1-step & 1 & 11.67 & 0.320 & \\
        \bottomrule
    \end{tabular}
    }
    \caption{Quantitative evaluations on zero-shot MS-COCO 2014 validation set (30K).}
    \label{tab:quantitative_cmp}
\end{table}

\begin{table}[!t]
    \centering
    \scalebox{0.9}{
    \begin{tabular}{l|cccc|c}
    \hline
    Models & Anim. & Concept & Painting & Photo	& Avg  \\ \hline
    SD-v2.0 (50 steps) & 27.48	& 26.89	& 26.86	& 27.46	& 27.17 \\
    SD-v1.4 (50 steps) & 27.26 & 26.61 & 26.66 & 27.27 & 26.95 \\ \midrule
    MD (50 steps) & 27.52 & 27.13 & 27.30 & 27.26 & \textbf{27.30} \\
    MD (1 step) & 27.05 & 26.33  & 26.41 & 26.80 & 26.65 \\
    \hline
    \end{tabular}
    }
    \caption{\scriptsize{HPS v2 benchmark.}}
    \label{tab:hpsv2}
    \vspace{-20pt}
\end{table}

\vspace{2mm} \noindent \textbf{On-device Benchmark} We conducted benchmarking of our proposed MD using tools\footnote{\url{https://github.com/apple/ml-stable-diffusion/tree/main.}} on iPhone 15 Pro. As depicted in Table~\ref{tab:model_parts}, MD exhibits superior efficiency in various aspects, including the text encoder, VAE decoder, UNet per-step inference, and the resulting overall latency.
\begin{table}[!ht]
    \centering
    \scalebox{0.85}{
    \begin{tabular}{lccccc}
    \toprule
        \textbf{Models}  &  \textbf{Text Encoder} & \textbf{Decoder} & \textbf{UNet} & \textbf{Steps} & \textbf{Overall} \\ \midrule
        SD 1.5~\cite{rombach2022high} & 4 & 285 & 357 & 20 & 7429 \\
        SnapFusion~\cite{li2023snapfusion} & 4 & 112 & 203 & 8 & 1740  \\
        UFOGen~\cite{ufogen2023} & 4 & 285 & 357 & 1 & 646 \\ \midrule
        \textbf{MD}-UFO & 4 & 92 & 142 &1 & 238 \\ 
        \bottomrule
    \end{tabular}
    }
    \caption{On-device latency (ms) measurements.}
    \label{tab:model_parts}
\end{table}

\vspace{-10mm}
\subsection{Ablation on UFOGen design choices} \label{sec:exp_ablate}
In this section, we ablate the design choices of UFOGen discussed in Section \ref{sec:method_ufo}. In particular, we tried different formulations for reconstruction loss and compared LoRA fine-tuning versus full parameter fine-tuning. We report the FID scores in Table \ref{tab:ufo_abalation}. Our ablation results indicate that
\begin{itemize}
    \item LoRA adversarial fine-tuning consistently performs worse than full parameter fine-tuning
    \item All reconstruction losses are effective in stabilizing the adversarial fine-tuning. The distill loss obtains slightly better performance.
\end{itemize}
\begin{table}[!ht]
    \centering
    \scalebox{0.95}{
    \begin{tabular}{l|cc}
    \toprule
        Reconstruction loss & No LoRA & LoRA \\ \midrule
        Diffusion loss& 11.67& 13.45\\
        Distill loss &11.20 & 13.24\\
        EMA distill loss & 12.08 & 14.05\\
        \bottomrule
    \end{tabular}}
    \caption{Ablation on different adversarial finetuning settings.}
    \label{tab:ufo_abalation}
\end{table}
Despite the slight advantage of using distill loss reflected in the FID metric, we conduct more in-depth study on the impact of training design choices on downstream application. Due to the constraint of space, we provide discussion on the study in the Appendix. We observe that, surprisingly, the original diffusion loss has the best results on applications. Therefore, our comprehensive exploration of the design space of UFOGen's training suggest that the original setting, namely full parameter fine-tuning with diffusion loss, is the best recipe for one-step diffusion model. As a result, we adopt this setting throughout the paper. 
\subsection{Applications} \label{sec:application}
\begin{figure}[!ht]
    \centering
    \includegraphics[width=0.7\textwidth]{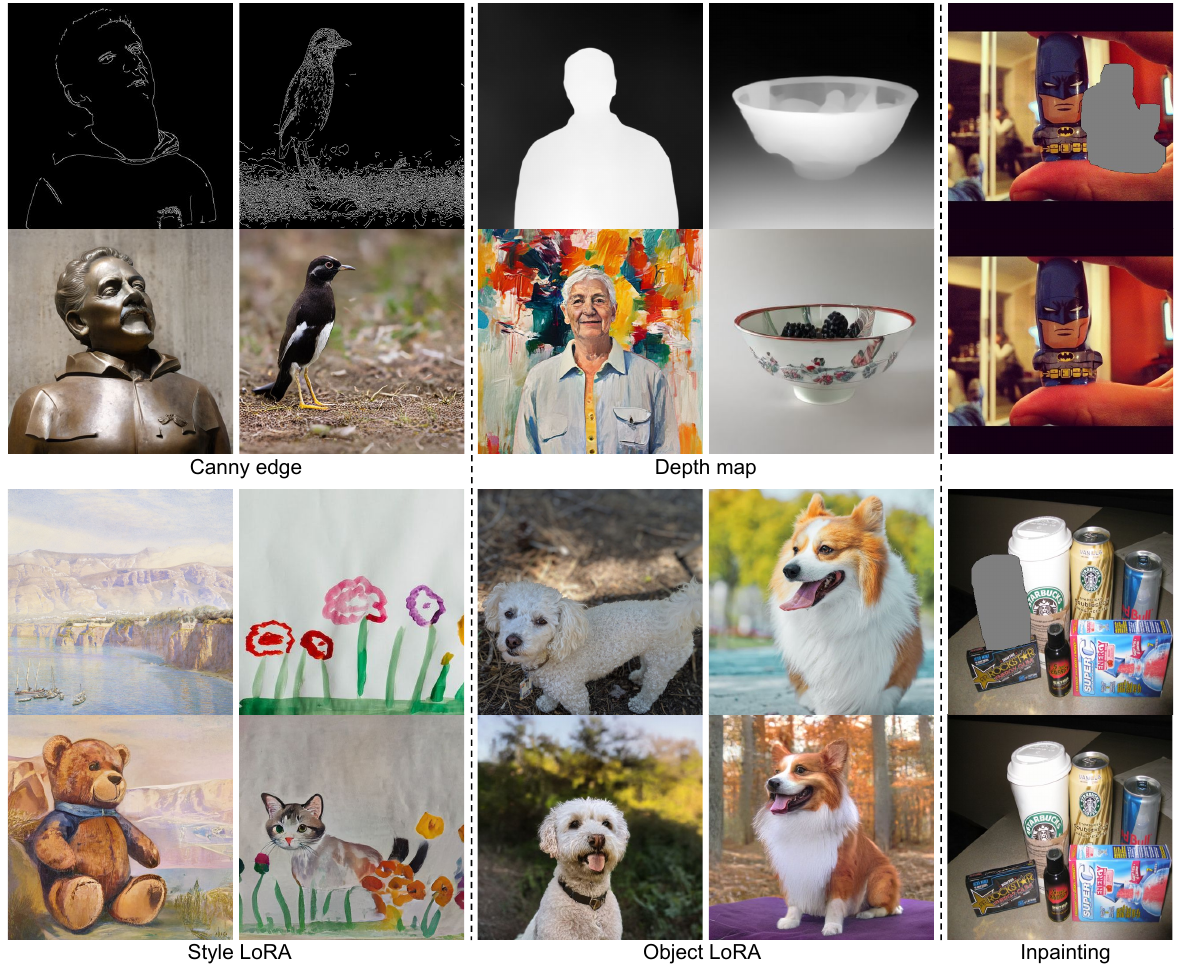}
    \caption{MD applications on plugins, LoRA finetuning and inpainting. In each group, top row illustrates the condition or reference images and bottom row shows the results.}
    \vspace{-10mm}
    \label{fig:main-application}
\end{figure}

Our MobileDiffusion framework supports a wide range of downstream conditional generation applications, as illustrated in Figure \ref{fig:main-application}. On tasks including controllable generation, personalized generation and in-painting, our model consistently generate visually appealing results instantly. We want to emphasize that the conditional generation models are trained on the pre-trained diffusion models, and they are seamlessly adapted to the one-step diffusion model without any re-training. Such a flexibility enables our model to be compatible with more potential applications.  
\section{Conclusions}\label{sec:conclusion}
In this paper, we explore pushing the boundary of the diffusion model's efficiency by proposing MobileDiffusion, with the ultimate goal of democratizing text-to-image generation on mobile devices. To achieve this goal, we conduct comprehensive studies of the architecture optimization for text-to-image diffusion models, which is a rarely touched area in prior work. Through the studies, we obtained a highly optimized architecture for diffusion UNet, with less than 400 million parameters and more efficient computational operations, while maintaining the quality. Additionally, we study the adversarial fine-tuning techniques and provide the best recipe for one-step diffusion sampling that excels in both text-to-image generation and various downstream applications. With the combined efforts, we are able to achieve the astonishing inference time of 0.2 seconds on mobile devices. 

\newpage

%
%
\bibliographystyle{splncs04}
\bibliography{main}

\appendix
\label{sec appendix}

\section{Architecture Optimization}
Here, we provide additional complementary details that were not covered in the primary section on architecture optimization. These details are intended to offer a more comprehensive understanding of the importance and necessity of certain components.

\paragraph{Replace $\mathsf{gelu}$ with $\mathsf{swish}$.}
As observed in Eq.~\ref{eq:gelu}, the function \textsf{gelu} contains a cubic term that may lead to error accumulation or overflow concerns, particularly when implementing quantization. In contrast, \textsf{swish} as shown in Eq.~\ref{eq:swish} is computationally simpler, offering a solution to mitigate potential problems.

\begin{equation}\label{eq:gelu}
\begin{split}
    \mathsf{gelu}(x)  & = x \cdot \Phi (x) \\
    & \approx 0.5x(1 + \mathsf{tanh}(\sqrt{\frac{2}{\pi}}(x+0.044715x^3)))
\end{split}
\end{equation}
where $\Phi(\cdot)$ is the cumulative distribution function for Gaussian distribution. 
\begin{equation}\label{eq:swish}
    \mathsf{swish}(x) = x \cdot \mathsf{sigmoid}(x)
\end{equation}

\begin{figure}[!ht]
    \centering
    \begin{tabular}{cc}
        \includegraphics[width=0.25\linewidth]{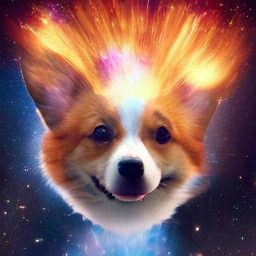} &
        \includegraphics[width=0.25\linewidth]{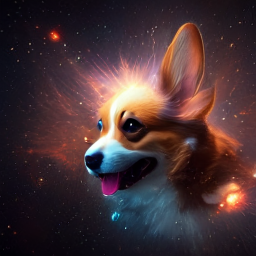} \\
        \multicolumn{2}{c}{\footnotesize{\textit{A corgi’s head depicted as an explosion of a nebula.}}} \\
    \end{tabular}
    \caption{Visual comparison between using $\mathsf{softmax}$ (left) and $\mathsf{relu}$ (right).}
    \label{fig:relu_vs_softmax}
\end{figure}

\paragraph{Finetune $\mathsf{softmax}$ into $\mathsf{relu}$.}
Figure~\ref{fig:relu_vs_softmax} presents a comparison of results obtained from models trained using the two activation functions. In our observations, there is rarely any noticeable visual distinction when using either of these activations in attention computations. However, employing the $\mathsf{relu}$ activation can significantly enhance mobile efficiency. As far as our knowledge extends, this represents the inaugural effort to effectively fine-tune a generative model initially activated with $\mathsf{softmax}$ into its $\mathsf{relu}$ counterpart.

\section{Mobile Diffusion Architectures}
In the main part, we have explained how we build efficient MD models step by step. Here we will show the detailed architectures of both MD and MD-lite in Table~\ref{tab:archs}.
\begin{table}[!h]
    \centering
    \scalebox{0.85}{
    \begin{tabular}{c|l|cc}
    \toprule
        \multirow{2}{*}{Models} & \multirow{2}{*}{Blocks} &  \multicolumn{2}{c}{Layers} \\
        & & Conv & Transformer \\ \midrule
        \multirow{5}{*}{MD} & Down-1 & FullConv $\times$1 & - \\
        & Down-2 & FullConv  $\times$1& CA \\
        & Down-3 & SPConv $\times$2 & (SA $\times$3 + CA $\times$3)$\times$2  \\
        & Up-3 & SPConv  $\times$3&  (SA $\times$3 + CA $\times$3) $\times$3 \\
        & Up-2 & FullConv  $\times$2& CA $\times$2 \\
        & Up-1 & FullConv $\times$2 & - \\ \midrule
        \multirow{5}{*}{MD-Lite} & Down-1 & FullConv  $\times$1& -\\
        & Down-2 & FullConv $\times$1 &  CA \\
        & Down-3 & SPConv  $\times$2& (SA $\times$3 + CA $\times$3) $\times$2 \\
        & Up-3 & SPConv  $\times$3& (SA $\times$2 + CA $\times$2) $\times$3 \\
        & Up-2 & FullConv  $\times$2& CA $\times$2 \\
        & Up-1 & FullConv $\times$2 & -\\
        \bottomrule
    \end{tabular}
    }
    \caption{Arcitecture details of MD and MD-Lite. FullConv means original residual blocks while SPConv uses Figure~\ref{fig:unet}. A notable difference between MD and MD-Lite is innermost channels which are 1024 and 896 respectively.}
    \label{tab:archs}
\end{table}

\section{Extended Generated Examples}
In this section, we present extended qualitative results of text to image, adapters and LoRA finetuing.

\subsection{Qualitative Examples}
\subsection{Text to Image}
\begin{figure*}[!h]
    \centering
    \includegraphics[width=0.95\textwidth]{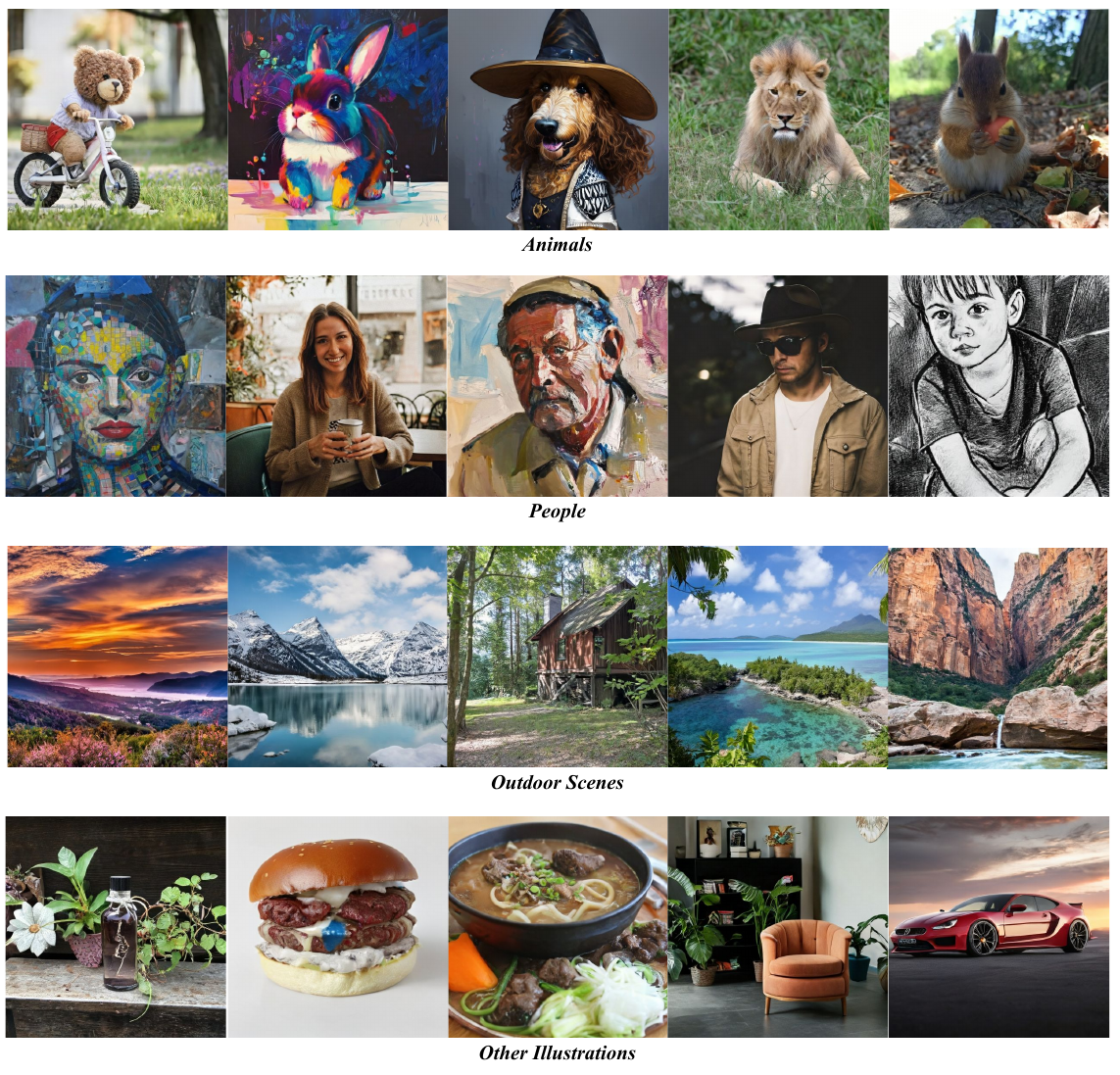}
    \caption{Extended one-step samples from MD.}
    \label{fig:extended_samples_1step}
\end{figure*}

In Figure \ref{fig:extended_samples_1step}, we select a wide range of  prompts to encompass various topics, including style, imagination, and complex compositions, beyond what was originally covered in the main section. This extended comparison underscores the consistent ability of MD to generate visually appealing results in just one-step with a highly efficient architecture. 

\subsection{Adapters and LoRA Finetuing}
\begin{figure*}[!h]
    \centering
    \includegraphics[width=0.95\textwidth]{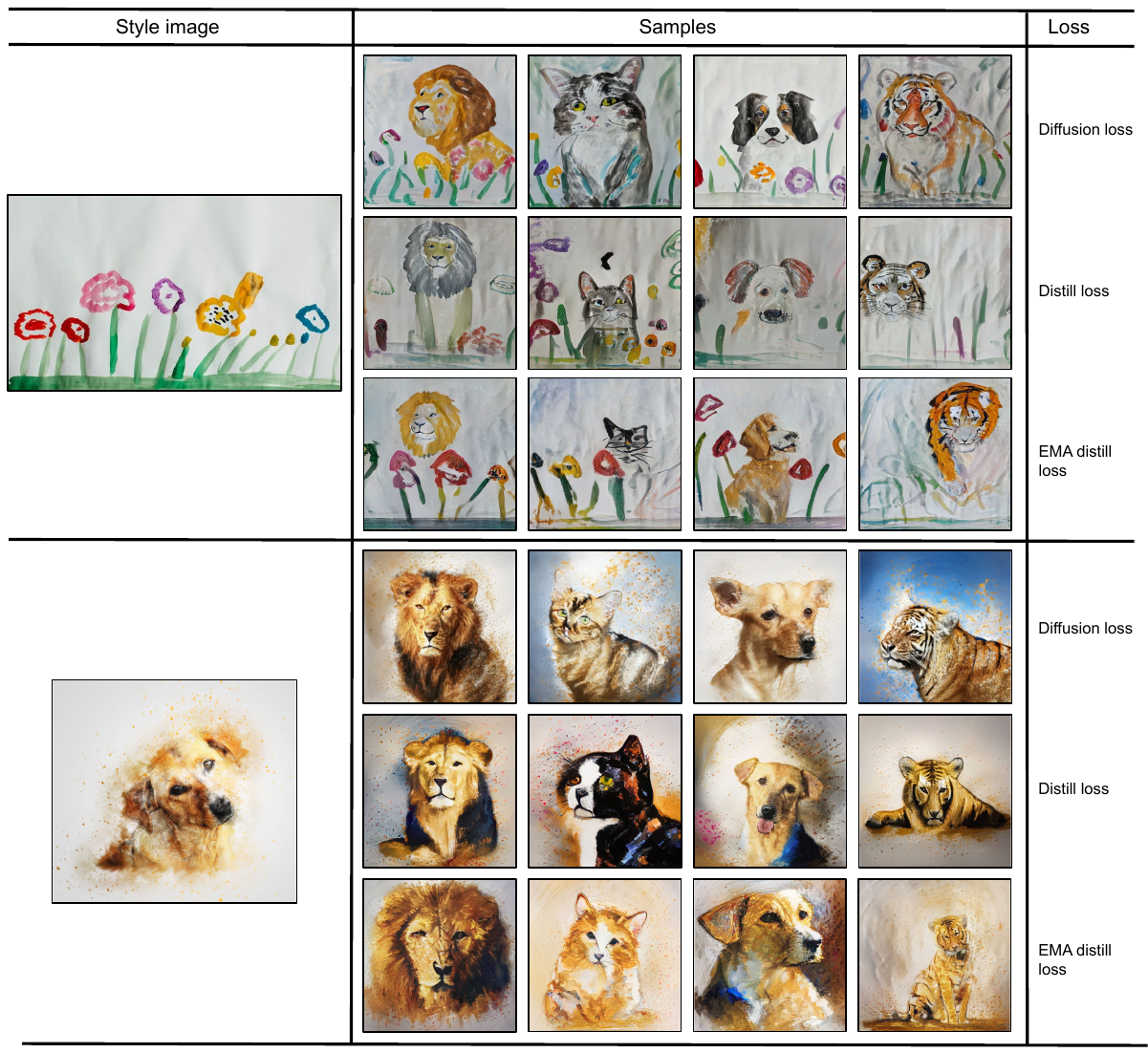}
    \caption{Ablation on different adversarial finetuning settings on LoRA applications. }
    \label{fig:extended_applications_lora}
\end{figure*}

\begin{figure*}[!h]
    \centering
    \includegraphics[width=0.95\textwidth]{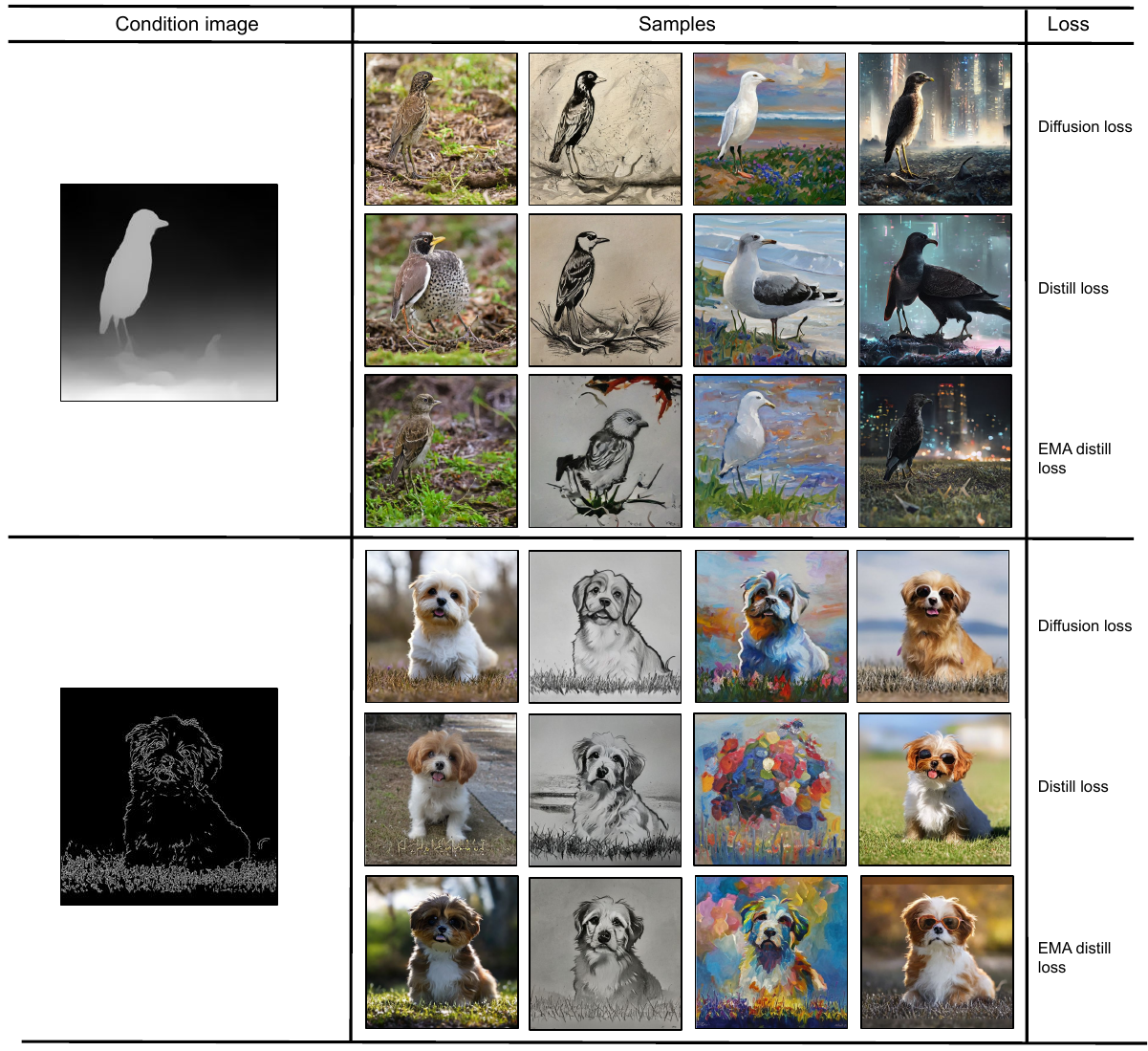}
    \caption{Ablation on different adversarial finetuning settings on adapter applications. }
    \label{fig:extended_applications_plugin}
\end{figure*}

We can seamlessly adapt pretrained adapters and LoRA weights to one-step diffusion models. In Figure~\ref{fig:extended_applications_plugin} and Figure~\ref{fig:extended_applications_lora}, we compare qualitative results from different settings. It's noteworthy that original UFOGen diffusion loss outperforms the other two in terms of condition and style faithfulness.

\begin{figure*}[!ht]
    \centering
    \includegraphics[width=0.95\textwidth]{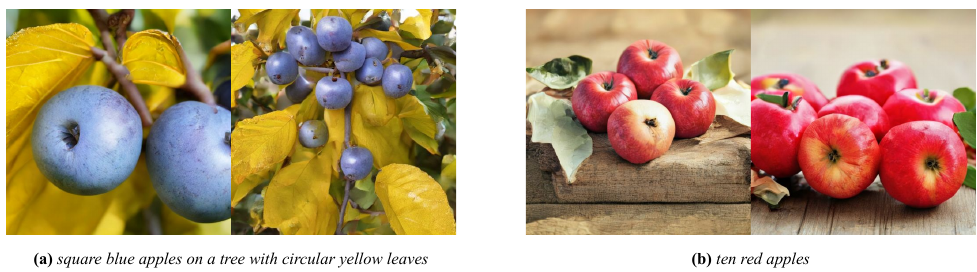}
    \caption{Failure cases of MD.}
    \label{fig:failure_cases}
\end{figure*}

\subsection{Failures Cases}
In Figure~\ref{fig:failure_cases}, we illustrate two prevalent types of failures: uncommon knowledge and quantity interpretation. We attribute these failures to the limited capabilities of the text encoder, as we have observed similar phenomena in both SD-1.5 and SD-XL.

\end{document}